\documentclass{article}
\usepackage{spconf,amsmath,graphicx}
\usepackage{enumerate}
\usepackage{graphicx}
\usepackage{amsmath,epsfig,amssymb,mathrsfs}
\usepackage{algorithmic}
\usepackage{cite}
\usepackage{amsthm}
\usepackage{arydshln}
\usepackage{amsopn}
\usepackage{color}
\usepackage{verbatim}
\usepackage{amsopn}
\usepackage{amssymb}
\usepackage{algorithm}
\usepackage{enumerate}
\usepackage{subcaption}
\usepackage{setspace} 
\usepackage{multirow}
\usepackage{float}
\usepackage{afterpage}
\usepackage{enumitem}
\usepackage{booktabs}
\usepackage{flushend}
\usepackage{lipsum}

\newcommand{\bx}{\mathbf{x}}

\newcommand{\bPhi}{\boldsymbol \Phi}
\title{Learning to Invert:  \\Signal Recovery via Deep Convolutional Networks}

\name{Ali Mousavi and Richard G. Baraniuk  \vspace{-2ex} 
\thanks{
Email: \{ali.mousavi, richb\} @rice.edu }}
\address{\\Department of Electrical and Computer Engineering \\
Rice University\\
Houston, TX 77005 \\}

\begin{document}
\maketitle


\begin{abstract}
The promise of compressive sensing (CS) has been offset by two significant challenges.  
First, real-world data is not exactly sparse in a fixed basis.
Second, current high-performance recovery algorithms are slow to converge, which limits CS to either non-real-time applications or scenarios where massive back-end computing is available.  
In this paper, we attack both of these challenges head-on by developing a new signal recovery framework we call {\em DeepInverse} that learns the inverse transformation from measurement vectors to signals using a {\em deep convolutional network}.  
When trained on a set of representative images, the network learns both a representation for the signals (addressing challenge one) and an inverse map approximating a greedy or convex recovery algorithm (addressing challenge two).
Our experiments indicate that the DeepInverse network closely approximates the solution produced by state-of-the-art CS recovery algorithms yet is hundreds of times faster in run time.
The tradeoff for the ultrafast run time is a computationally intensive, off-line training procedure typical to deep networks.  
However, the training needs to be completed only once, which makes the approach attractive for a host of sparse recovery problems.  
\end{abstract}



\begin{keywords}
Deep Learning, Compressive Sensing, Convolutional Neural Networks \vspace{-1.5ex}
\end{keywords}


\section{Introduction}
\label{sec:intro}


 An inverse problem that has many important applications is recovering $\mathbf{x} \in \mathbb{R}^N$ from a set of undersampled linear measurements $\mathbf{y} = \boldsymbol \Phi \mathbf{x} \in \mathbb{R}^M$, where $\boldsymbol \Phi$ is an $M\times N$ {\em measurement matrix} and $M \ll N$. 
This problem is ill-posed in general and hence, in order to successfully recover the original signal $\mathbf{x}$, it should have some type of structure such that its dimensionality can be reduced without losing information. 

Compressive sensing (CS) \cite{donoho2006compressed, baraniuk2007compressive, candes2006compressive} is a special case of this problem in which the signal has a sparse representation, i.e., there exists an $N\times N$ basis matrix $\boldsymbol \Psi = [\boldsymbol \psi_1|\boldsymbol \psi_2|\ldots|\boldsymbol \psi_N]$ such that $\mathbf{x} = \boldsymbol \Psi \mathbf{s}$ and only $K \ll N$ of the coefficients $\mathbf{s}$ are nonzero.
Recovering the signal $\mathbf{x}$ from the measurements $\mathbf{y}$ is effected by a sparsity-regularized convex optimization or greedy algorithm \cite{candes2006near, donoho2009message, needell2009cosamp}. 


The promise of CS has been offset by two significant challenges.  
The first challenge is that real-world data is not exactly sparse in a fixed basis.
Some work has been pursued on learning data-dependent dictionaries to sparsify signals \cite{mallat1999wavelet, duarte2008learning, aharon2006img}, but the redundancy of the resulting approaches degrades recovery performance.
The second challenge is that current high-performance recovery algorithms (e.g., \cite{metzler2014denoising}) are slow to converge, which limits CS to either non-real-time applications or scenarios where massive back-end computing is available.  

In this paper, we attack both of these challenges head-on by developing a new signal recovery framework we call {\em DeepInverse} that learns the inverse transformation from measurement vectors $\mathbf{y}$ to signals $\mathbf{x}$ using a {\em deep convolutional network}.  
When trained on a set of representative images, the network learns both a representation for the signals $\mathbf{x}$ (addressing challenge one) and an inverse map approximating a greedy or convex recovery algorithm (addressing challenge two).


Our experiments below indicate that the DeepInverse network closely approximates the solution produced by state-of-the-art CS recovery algorithms yet is hundreds of times faster in run time.
The tradeoff for the ultrafast run time is a computationally intensive, off-line training procedure typical to deep networks.  
However, the training needs to be completed only once, which makes the approach attractive for a host of sparse recovery problems.  

\vspace{-2ex} 
\section{Prior Work \vspace{-1.5ex}}

The first paper to study the problem of structured signal recovery from a set of undersampled measurements using a deep learning approach was \cite{mousavi2015deep}. 
This work employed a stacked denoising autoencoder (SDA) as an unsupervised feature learner. 
The main drawback of the SDA approach is that its network consists of fully-connected layers, meaning that all units in two consecutive layers are connected to each other. 
Thus, as the signal size grows, so does the network. 
This imposes a large computational complexity on the training (backpropagation) algorithm and can also lead to overfitting.
The solution to this issue that was implemented in \cite{mousavi2015deep} is to divide the signal into smaller non-overlapping or overlapping blocks and then sense/reconstruct each block separately. 
While such an approach deals with the curse of dimensionality, a blocky measurement matrix $\boldsymbol \Phi$ is unrealistic in many applications.

The above work was followed by \cite{kulkarni2016reconnet}, who used a fully-connected layer along with convolutional neural networks (CNNs) to recover signals from compressive measurements.
Their approach also used a blocky measurement matrix $\boldsymbol \Phi$.


In contrast to both \cite{mousavi2015deep} and \cite{kulkarni2016reconnet}, DeepInverse works with arbitrary (and not just blocky) measurement matrices $\boldsymbol \Phi$.




\section{Deep Convolutional Networks Primer}
\label{sec:CNN}





Deep Convolutional Networks (DCNs) consist of three major layers: first, convolutional layer that is the core of these networks. This layer consists of a set of learnable filters with a limited receptive field that are replicated across the entire visual field and form feature maps. Second, ReLU nonlinearity layer that causes nonlinearity in decision function of the overall network. Third, pooling layer that is a form of down-sampling and provides translation invariance. Backpropagation algorithm is used to train the whole network and fine tune filters of convolutional layers.   
All three layers play important roles in DCNs' primary application which is image classification. In other words, in DCNs one reduces dimensionality of a given image by a series of convolutional and pooling layers in order to extract the label of that image. Authors in \cite{patel2015probabilistic} have introduced a probabilistic framework that provides insights into the success of DCNs in image classification task.

DCNs have two distinctive features that make them uniquely applicable to sparse recovery problems. First, sparse connectivity of neurons. Second, having shared weights across the entire receptive fields of one layer which increases learning speed comparing to fully-connected networks.


\section{Convolutional Networks for \\ Signal Recovery}
\label{sec:CNN}
 \vspace{-1ex}

We now develop our new DeepInverse signal recovery framework that learns the inverse transformation from measurement vectors $\mathbf{y}$ to signals $\mathbf{x}$ using a special DCN.  
When trained on a set of representative images, the network learns both a representation for the signals $\mathbf{x}$ and an inverse map approximating a greedy or convex recovery algorithm.

\begin{figure*}[t!]
\begin{center}
\includegraphics[width= 17cm]{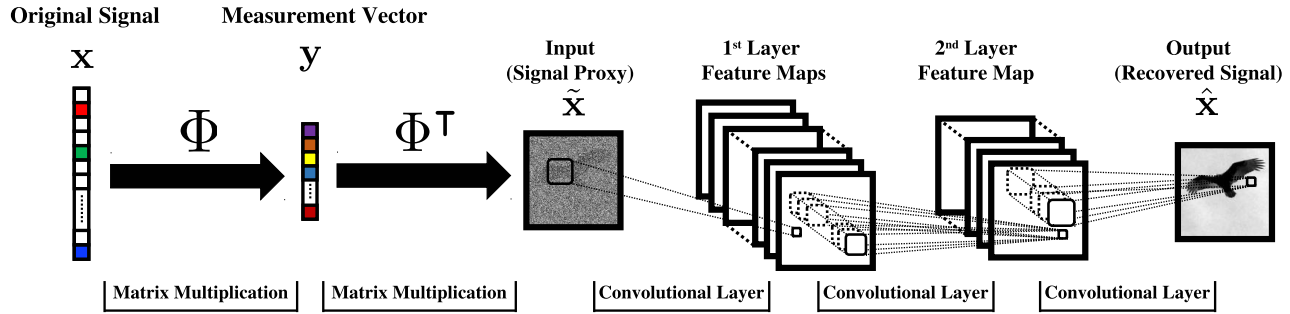}
\caption{
{\em DeepInverse} learns the inverse transformation from measurement vectors $\mathbf{y}$ to signals $\mathbf{x}$ using a special deep convolutional network.
\vspace{-3ex}}
\label{fig:proxy}
\end{center}
\end{figure*}

DeepInverse takes an input (set of measurements $\mathbf{y}$) in $\mathbb{R}^M$ and produces an output (signal estimate $\widehat{\mathbf{x}}$) in $\mathbb{R}^N$, where typically $M<N$.
Accomplishing this dimensionality increase requires several modifications to the conventional DCN architecture.
(See Fig.\ \ref{fig:proxy}.)
First, to boost the dimensionality of the input from $\mathbb{R}^M$ to $\mathbb{R}^N$, we employ a fully connected linear layer.
While, in general, one could learn the weights in this layer, in this paper, we set the weights to implement the adjoint operator $\boldsymbol \Phi^{\intercal}$.
Second, to preserve the dimensionality of the processing in $\mathbb{R}^N$, we dispense with the downsampling max-pooling operations.

As in the usual CS paradigm, we assume that the measurement matrix $\bPhi$ is fixed. 
Therefore, each $\mathbf{y}_i$ ($1 \leq i \leq M$) is a linear combination of $\mathbf{x}_j$s ($1\leq j \leq N$). The training set $\mathcal{D}_{\rm train} =\{(\mathbf{y}^{(1)},\mathbf{x}^{(1)}),(\mathbf{y}^{(2)},\mathbf{x}^{(2)}),\ldots, (\mathbf{y}^{(l)},\\ \mathbf{x}^{(l)})\}$ consists of $l$ pairs of signals and their corresponding measurements. Similarly, test set $\mathcal{D}_{\rm test} = \{(\mathbf{y}^{(1)},\mathbf{x}^{(1)}),(\mathbf{y}^{(2)},\\ \mathbf{x}^{(2)}), \ldots,(\mathbf{y}^{(s)},\mathbf{x}^{(s)})\}$ consists of $s$ pairs including original signals and their corresponding measurements. By training a DCN, we learn a nonlinear mapping from a signal proxy $\mathbf{\tilde{x}}$ to its original signal $\bx$.

In the experiments described below, we use one fixed fully connected layer (to implement $\boldsymbol \Phi^{\intercal}$) and three convolutional layers.
As in a DCN, each convolutional layer applies a ReLU nonlinearity to its output. 


A few details are in order. We denote signal proxy by $\mathbf{\tilde{x}}$ where $\mathbf{\tilde{x}}=\boldsymbol \Phi^{\intercal} \mathbf{y} $
We assume that $\mathbf{\tilde{x}}$ is $n_1 \times n_2$ (where $n_1\times n_2 = N$).
Then the $(i,j)$-th entry of the $k$-th feature map in the first convolutional layer receives $\mathbf{\tilde{x}}$ as its input; its output is given by
\begin{align}
(\bx_{c_1})_{i,j}^k=\mathcal{S}({\rm ReLU}((\mathbf{W_1^k} \ast \mathbf{\tilde{x}})_{i,j}+(\mathbf{b_1^k})_{i,j})), 
\end{align}
where $\mathbf{W_1^k} \in \mathbb{R}^{k_1 \times k_2}$ and $\mathbf{b_1^k} \in \mathbb{R}^{n_1+k_1-1 \times n_2+k_2-1}$ denote the filter and bias values corresponding to the $k$-th feature map of the first layer and ${\rm ReLU}(x) = \max(0,x)$. 
Finally, the subsampling operator $\mathcal{S}(\cdot)$ takes the output of ${\rm ReLU}(\cdot)$ to the original signal size by ignoring the borders created by zero-padding the input. 

The feature maps of the second and third convolutional layers are developed in a similar manner.
While the filter shapes and and biases could be different in the second and third layers of the network, the principles in these layers are the same as first layer. 
Let $\ell_1, \ell_2$, and $\ell_3$ denote the number of filters in the first, second, and third layers, respectively. 
If we denote the output of this convolutional network by $\mathbf{\hat{x}}$ and its set of parameters by $\Omega = \left\{ \{\mathbf{W_1^k},\mathbf{b_1^k}\}_{k=1}^{\ell_1}, \{\mathbf{W_2^k},\mathbf{b_2^k}\}_{k=1}^{\ell_2}, \{\mathbf{W_3^k},\mathbf{b_3^k}\}_{k=1}^{\ell_3} \right\}$, then we can define a nonlinear mapping from the measurements to original signal as $\mathbf{\hat{x}}=\mathcal{M}(\mathbf{y},\Omega)$.

\sloppy
Using the mean squared error (MSE) as a loss function over the training data $\mathcal{D}_{\rm train}$ defined as $\mathcal{L}(\Omega)=\frac{1}{l}\sum_{i=1}^l \| \mathcal{M}(\mathbf{y}^{(i)},\Omega)-\mathbf{x}^{(i)}\|_2^2$, we can employ backpropagation \cite{rumelhart1988learning} in order to minimize $\mathcal{L}(\Omega)$ and learn the parameters.
 \vspace{-2ex}



\section{Experimental Results}
\label{sec:simul}

In this section we describe the implementation of DeepInverse and compare its performance to several other state-of-the-art CS recovery algorithms.

As we mentioned earlier, one of the main goals of this paper is to show that we can use deep learning framework to recover images from undersampled measurements without any need to divide images into small blocks and recover each block separately. For this purpose, we use DeepInverse that receives a signal proxy, i.e., $\mathbf{\tilde{x}}=\bPhi^{\intercal} \mathbf{y}$ (with same dimension as $\bx$) as its input. In addition, it has 3 layers with the following specifications.
The first layer has 64 filters, each having 1 channel of size $11 \times 11$. 
The second layer has 32 filters, each having 64 channels of size $11 \times 11$. 
The third layer has 1 filter with 32 channels of size $11 \times 11$. 
We trained DeepInverse using $64 \times 64$ cropped subimages of the natural images in the {\em ImageNet} dataset \cite{russakovsky2014imagenet}.
Test images were drawn from ImageNet images that were not used for training purposes.

\begin{figure}[H]
\begin{center}
\includegraphics[width= 7cm]{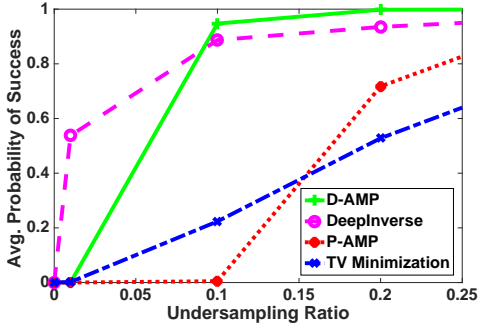}
\caption{Average probability of successful signal recovery for
different sampling rates and algorithms.}
\label{fig:phasetransition}
\end{center}
\end{figure}

Figure \ref{fig:phasetransition} shows the plot of average probability of successful recovery for different undersampling ratios ($M/N$) and three different recovery algorithms:
D-AMP \cite{metzler2014denoising}, total variation (TV) minimization \cite{candes2006robust}, and P-AMP \cite{mousavi2015consistent}. 
Note that we do not include any results from \cite{mousavi2015deep,kulkarni2016reconnet} in our simulation results, since these approaches are specifically designed for block-based recovery whereas in this paper we focus on recovering signals without subdivision.

Figure \ref{fig:phasetransition} compares the probability of successful recovery as measured by 2000 Monte Carlo samples. 
For each under-sampling ratio and Monte Carlo sample, we define the success variable $\phi_{\delta,j}=\mathbb{I}\left(\frac{\| \hat{\bx}^{(j)}-\bx^{(j)} \|_2^2}{\| \bx^{(j)} \|_2^2} \le 0.1 \right)$. 
For small values of undersampling ratio (e.g., 0.01) DeepInverse has better performance than state-of-the-art recovery methods. 
However, as the undersampling ratio increases, D-AMP outperforms DeepInverse. 
Although Figure \ref{fig:phasetransition} shows that for every undersampling ratio one method works better than others, there is not a clear winner in terms of reconstruction quality. 

\begin{figure}[H]
\begin{center}
\includegraphics[width= 7cm]{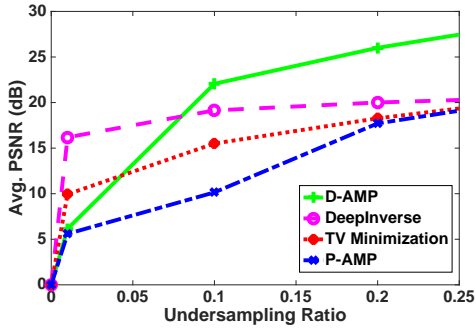}
\caption{ Average PSNR (dB) for different sampling rates and algorithms.}
\label{fig:PSNR}
\end{center}
\end{figure}

\begin{figure}[H]
\begin{center}
\includegraphics[width= 5.5cm]{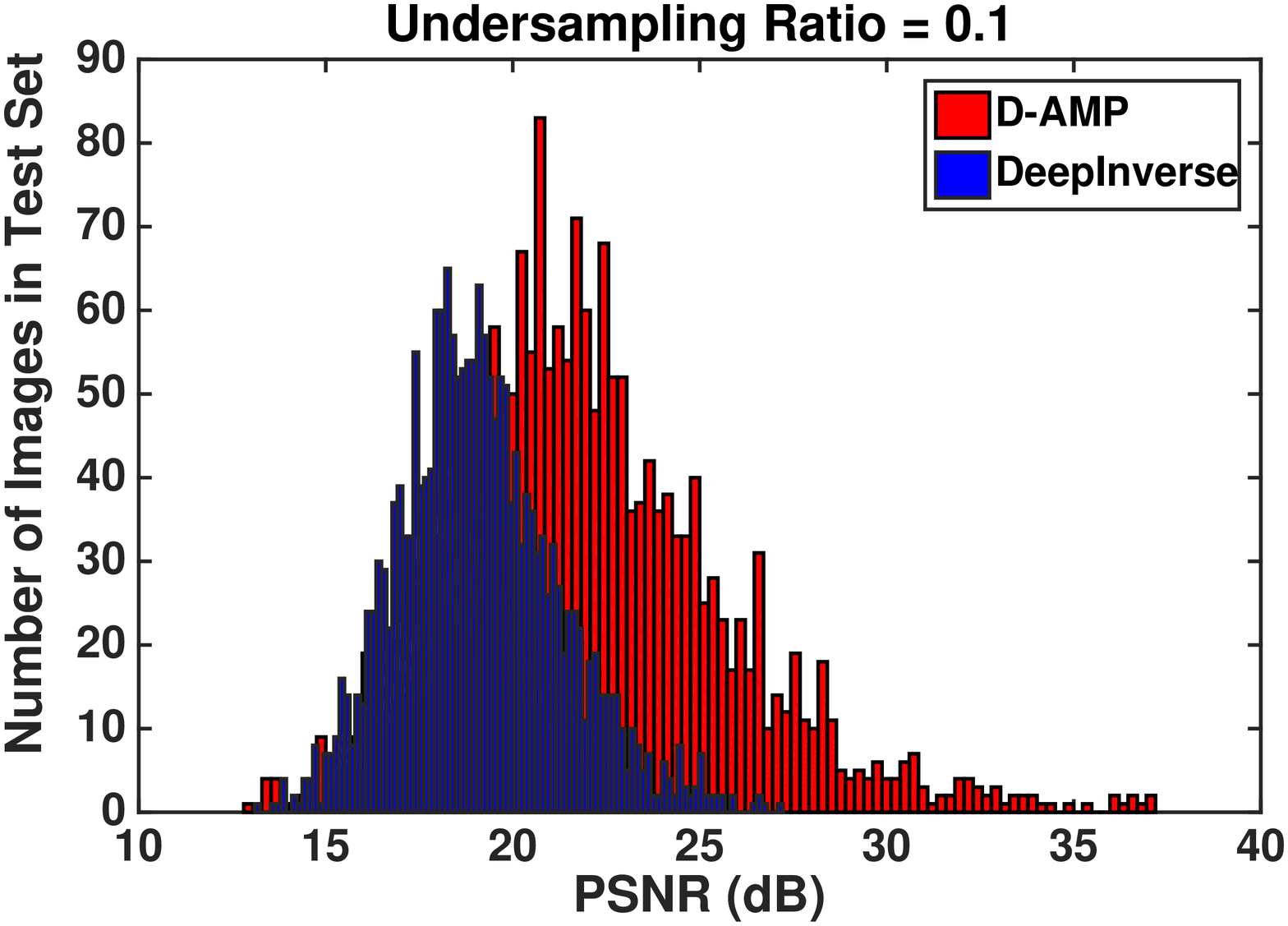}
\includegraphics[width= 5.5cm]{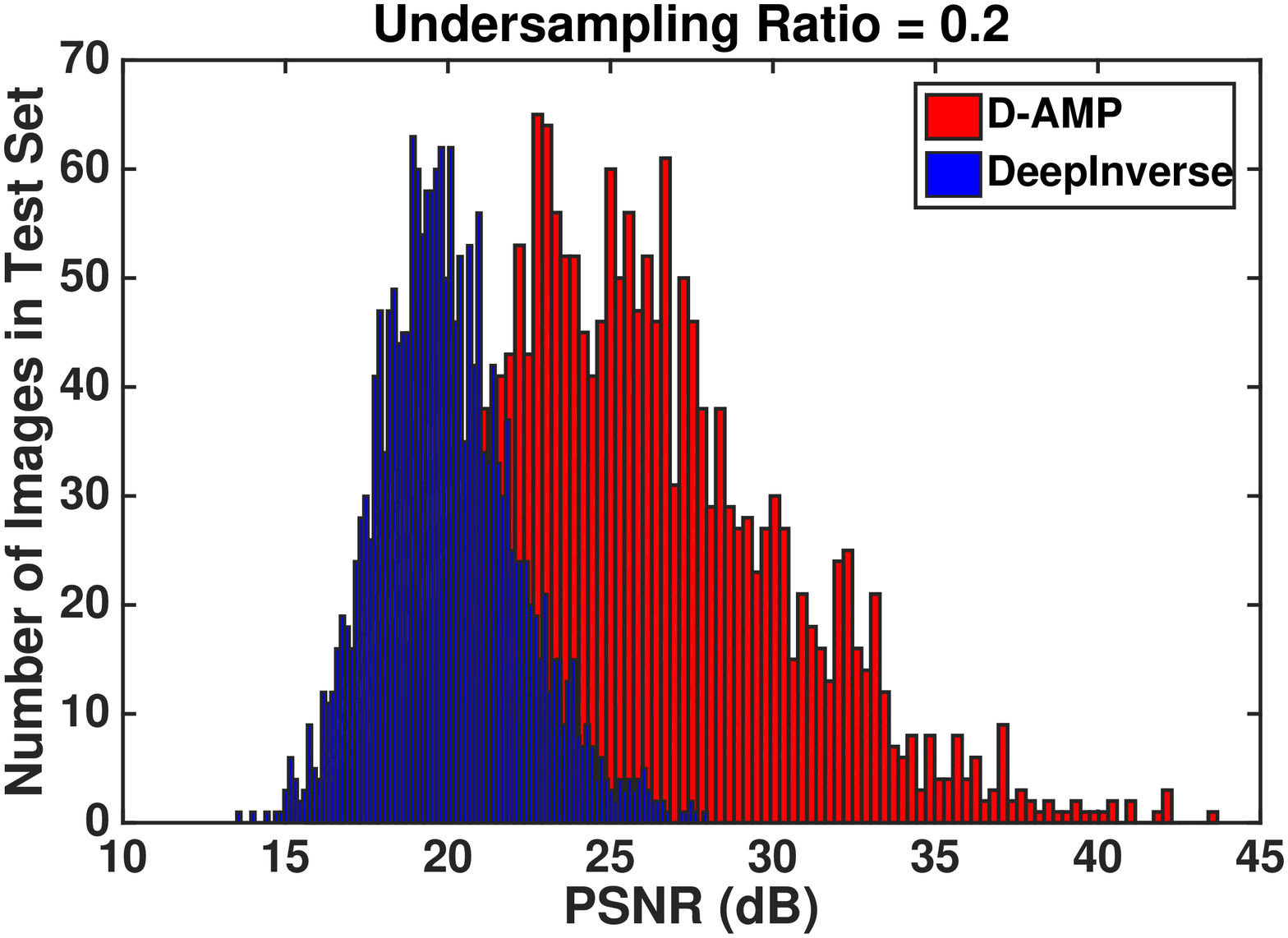}
\caption{Histograms of the PSNRs of the test images for two different sampling rates as recovered by D-AMP and DeepInverse.}
\label{fig:hist}
\end{center}
\end{figure}

Figure \ref{fig:PSNR} compares the average PSNR \footnote{${\rm PSNR} = 10.\log_{10}\left(\frac{\max_{{\rm Image}}^2}{{\rm MSE}}\right)$} of the Monte Carlo test samples for different undersampling ratios and algorithms.
Figure \ref{fig:hist} shows the histograms of the PSNRs of the recovered test images, indicating the DeepInverse outperforms D-AMP for some images in the test set.

While Figs.\ \ref{fig:phasetransition} and \ref{fig:PSNR} indicate that DeepInverse offers recovery probability and PSNR performance that is comparable to state-of-the-art CS recovery algorithms, Table \ref{table:time} shows that DeepInverse has a run time that is a tiny fraction of current algorithms.  
This fact makes DeepInverse especially suitable for applications that need low-latency recovery.

Table \ref{table:results} plots the images recovered by DeepInverse and D-AMP when they are on their best and worst behavior.

Table \ref{table:noise} shows the effect of adding input noise on recovery performance of D-AMP and DeepInverse. We can see that for undersampling ratio of 0.1 and 20 dB input noise, DeepInverse is more robust to noise comparing to D-AMP.

\begin{figure}[H]
\begin{center}
\includegraphics[width= 7cm]{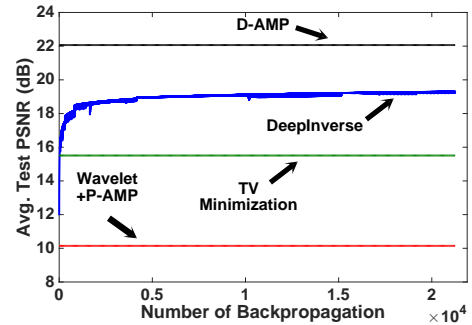}
\caption{Convergence of backpropagation training over different iterations for DeepInverse for the undersampling ratio $M/N=0.1$.}
\label{fig:bp_converge}
\end{center}
\end{figure}

Finally, Figure \ref{fig:bp_converge} shows the convergence of the back-propagation training algorithm over different iterations for DeepInverse. It also shows the average PSNR of the images in the test dataset for different methods with $M/N=0.1$. 
We can see that after several iterations DeepInverse starts to outperform TV minimization and P-AMP.


Although D-AMP has better performance than a 3-layer DeepInverse in general, we should consider two points. First, by training an DeepInverse with more layers the network will have a larger capacity and hence, we expect it to offer better recovery performance. 
We leave studying DeepInverses with larger capacities as a topic of our future work. 
Second, DeepInverse is specially useful for applications that need low-latency recovery and at the same time we are not able to divide images into smaller blocks, i.e., we need to apply a sensing matrix to the entire signal rather than smaller blocks of it.

\begin{table}[H]
\centering
\caption{Average reconstruction time of test set images for different sampling rates and algorithms.}
\begin{tabular}{|c|c|c|l|c|}
\hline
\multicolumn{1}{|l|}{\multirow{2}{*}{$\frac{M}{N}$}} & \multicolumn{4}{c|}{Reconstruction Time (s)}                                             \\ \cline{2-5} 
\multicolumn{1}{|l|}{}                               & \multicolumn{1}{l|}{DeepInverse} & \multicolumn{1}{l|}{D-AMP} & TV   & \multicolumn{1}{l|}{P-AMP} \\ \hline
0.2                                                  & \textbf{0.01}                      & 3.41                       & 2.53 & 1.53                        \\ \hline
0.1                                                  & \textbf{0.01}                      & 2.93                       & 2.34 & 1.23                        \\ \hline
0.01                                                 & \textbf{0.01}                      & 2.56                       & 2.26 & 0.94                        \\ \hline
\end{tabular}
\label{table:time}
\end{table}

\if
delete this table
\begin{table}[H]
\centering
\caption{Reconstruction quality (PSNR) for best and worst recovered images from DeepInverse and D-AMP when the sampling rate is $M/N=0.1$. We have reported P-AMP and TV-Min PSNRs as well while omitting the corresponding images.}
\begin{tabular}{ l || l | l | l | l }
       & \includegraphics[width=32px, height=32px, trim=0 0 0 -5]{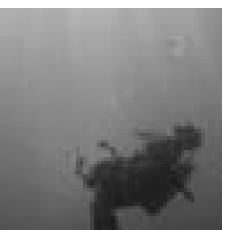} & \includegraphics[width=32px, height=32px, trim=0 0 0 -5]{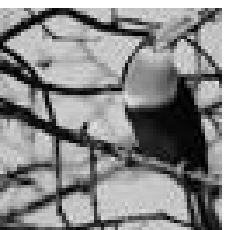} & \includegraphics[width=32px, height=32px, trim=0 0 0 -5]{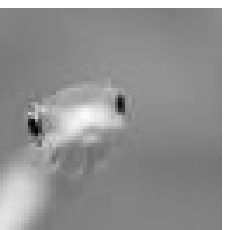} & \includegraphics[width=32px, height=32px, trim=0 0 0 -5]{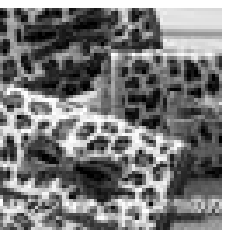} \\ \hline \hline
DeepInverse   & 27.23 & 13.10 &  26.54 & \textbf{13.55}  \\ \hline
D-AMP   & \textbf{33.60} & \textbf{13.19} & \textbf{37.18} & 13.40 \\ \hline
TV-Min &  22.24 & 6.28 & 26.32  & 9.23  \\ \hline
P-AMP  & 15.49 & 6.20  &  19.93 & 9.04 \\ 
\end{tabular}
\label{table:results}
\end{table}
end of delete
\fi

\begin{table}[H]
\centering
\caption{Quality of reconstruction (PSNR in dB) for best (denoted by $\uparrow$) and worst (denoted by $\downarrow$) reconstructed images by DeepInverse (DI) and D-AMP when undersampling ratio is 0.1.}
\label{table:results}
\begin{tabular}{c||l|l|l|l}
Condition & DAMP $\uparrow$& DAMP $\downarrow$ & DI $\uparrow$ & DI $\downarrow$\\ \hline \hline
Image     &   \includegraphics[width=32px, height=32px, trim=0 0 0 -5]{image632.eps}         &  \includegraphics[width=32px, height=32px, trim=0 0 0 -5]{image1309.eps}             &      \includegraphics[width=32px, height=32px, trim=0 0 0 -5]{image148.eps}          &        \includegraphics[width=32px, height=32px, trim=0 0 0 -5]{image336.eps}          \\ \hline  \hline

DeepInverse       &  \begin{tabular}[c]{@{}c@{}}\includegraphics[width=32px, height=32px, trim=0 0 0 -5]{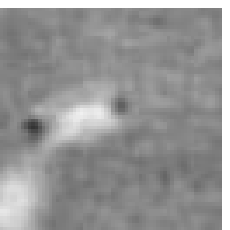}\\ 26.54\end{tabular}      &         \begin{tabular}[c]{@{}c@{}}\includegraphics[width=32px, height=32px, trim=0 0 0 -5]{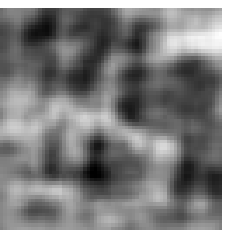}\\ \textbf{13.55}\end{tabular}          &      \begin{tabular}[c]{@{}c@{}}\includegraphics[width=32px, height=32px, trim=0 0 0 -5]{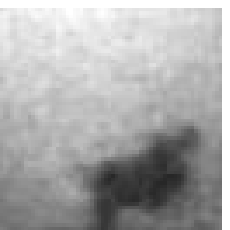}\\ 27.23\end{tabular}        &       \begin{tabular}[c]{@{}c@{}}\includegraphics[width=32px, height=32px, trim=0 0 0 -5]{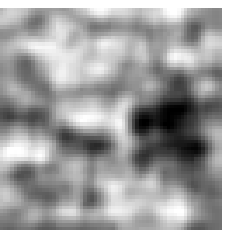}\\ 13.10\end{tabular}       \\ \hline
DAMP      &      \begin{tabular}[c]{@{}c@{}}\includegraphics[width=32px, height=32px, trim=0 0 0 -5]{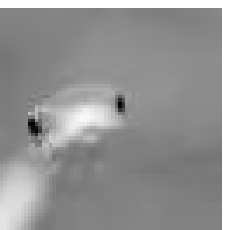}\\ \textbf{37.18}\end{tabular}              &          \begin{tabular}[c]{@{}c@{}}\includegraphics[width=32px, height=32px, trim=0 0 0 -5]{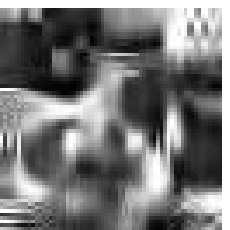}\\ 13.19\end{tabular}           &           \begin{tabular}[c]{@{}c@{}}\includegraphics[width=32px, height=32px, trim=0 0 0 -5]{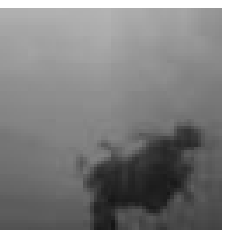}\\ \textbf{33.60}\end{tabular}         &          \begin{tabular}[c]{@{}c@{}}\includegraphics[width=32px, height=32px, trim=0 0 0 -5]{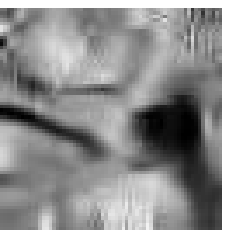}\\ \textbf{13.40}\end{tabular}    \\ 
\end{tabular}
\end{table}

\begin{table}[H]
\centering
\caption{Effect of noise on average PSNR (dB) of reconstructed test images for D-AMP and DeepInverse (undersampling ratio = 0.1). We added 20 dB noise to images of test set. Due to noise-folding \cite{davenport2012pros}  the variance of the noise that we observe after the reconstruction is larger than the input noise.}
\label{table:noise}
\begin{tabular}{l||l|l}
                          & \multicolumn{1}{c|}{\begin{tabular}[c]{@{}c@{}}Noiseless\\ Measurements\end{tabular}} & \multicolumn{1}{c}{\begin{tabular}[c]{@{}c@{}}Noisy \\ Measurements\end{tabular}} \\ \hline \hline
\quad DAMP  &                                             \qquad   22.06                                       &                  \qquad 21.14           \\ \hline
DeepInverse               &                              \qquad    19.14                                                      &          \qquad         18.70    \\ 
\end{tabular}
\end{table}

\vspace{-3ex}
\section{Conclusions}
\label{sec:conclu}
\vspace{-2ex}
In this paper, we have developed the DeepInverse framework for sensing and recovering signals. We have shown that this framework can learn a structured representation from training data and efficiently approximate a signal recovery at a small fraction of the cost of state-of-the-art recovery algorithms.


\bibliographystyle{IEEEbib}
\bibliography{paper}

\end{document}